\documentclass[runningheads]{llncs}
\usepackage[T1]{fontenc}
\usepackage{graphicx}
\usepackage{multirow}
\usepackage{pifont}
\usepackage{booktabs}
\usepackage{amsfonts}
\usepackage{dsfont}
\usepackage{amsmath}
\usepackage{wrapfig}
\usepackage{bbding}
\usepackage[table]{xcolor}

\begin{document}
\title{$R^2$-Mesh: Reinforcement Learning Powered Mesh Reconstruction via Geometry and Appearance Refinement}
\titlerunning{$R^2$-Mesh}
%
\author{Haoyang Wang \and
Liming Liu \and
Xinggong Zhang\inst{(}\Envelope\inst{)}}
\authorrunning{H. Wang et al.}
%
\institute{Peking University\\
\email{\{haoyang.wang,llm\}@stu.pku.edu.cn}\\
\email{zhangxg@pku.edu.cn}}
\maketitle              
\begin{abstract}
Mesh reconstruction from Neural Radiance Fields (NeRF) is widely used in 3D reconstruction and has been applied across numerous domains. However, existing methods typically rely solely on the given training set images, which restricts supervision to limited observations and makes it difficult to fully constrain geometry and appearance. Moreover, the contribution of each viewpoint for training is not uniform and changes dynamically during the optimization process, which can result in suboptimal guidance for both geometric refinement and rendering quality. To address these limitations, we propose $R^2$-Mesh, a reinforcement learning framework that combines NeRF-rendered pseudo-supervision with online viewpoint selection. Our key insight is to exploit NeRF’s rendering ability to synthesize additional high-quality images, enriching training with diverse viewpoint information. To ensure that supervision focuses on the most beneficial perspectives, we introduce a UCB-based strategy with a geometry-aware reward, which dynamically balances exploration and exploitation to identify informative viewpoints throughout training. Within this framework, we jointly optimize SDF geometry and view-dependent appearance under differentiable rendering, while periodically refining meshes to capture fine geometric details. Experiments demonstrate that our method achieves competitive results in both geometric accuracy and rendering quality.

\end{abstract}

\section{Introduction}

\label{sec:intro}

Mesh reconstruction of 3D scenes is profoundly essential in domains such as virtual reality \cite{mesh_vr}, medical imaging \cite{mesh_heart}, and robotics \cite{mesh_robot}. High-quality 3D meshes enable immersive interaction, precise analysis, and reliable perception. However, reconstructing meshes directly from RGB images remains a challenging problem. Real-world scenes often contain occlusions, non-uniform lighting, and complex textures. These factors, together with the need to capture fine-grained geometric structures, make it difficult to obtain faithful reconstructions from image observations.

The recent advent of Neural Radiance Fields (NeRF) \cite{kplanes,nerf,instantngp} has brought transformative progress to 3D reconstruction. NeRF excels at modeling volumetric density and view-dependent appearance, achieving photorealistic novel view synthesis. Motivated by this capability, numerous works \cite{neuralangelo,neus,neus2,nerfmeshing} employ NeRF as an intermediate representation for mesh extraction. Typical strategies either transform radiance fields into Signed Distance Fields (SDFs) followed by Marching Cubes \cite{marching-cubes}, or refine explicit meshes under differentiable rendering supervision \cite{mobilenerf,nerf2mesh}. 

\begin{wrapfigure}{r}{0.6\textwidth}
  \centering
  \vspace{-35pt}  
  \includegraphics[width=0.58\textwidth]{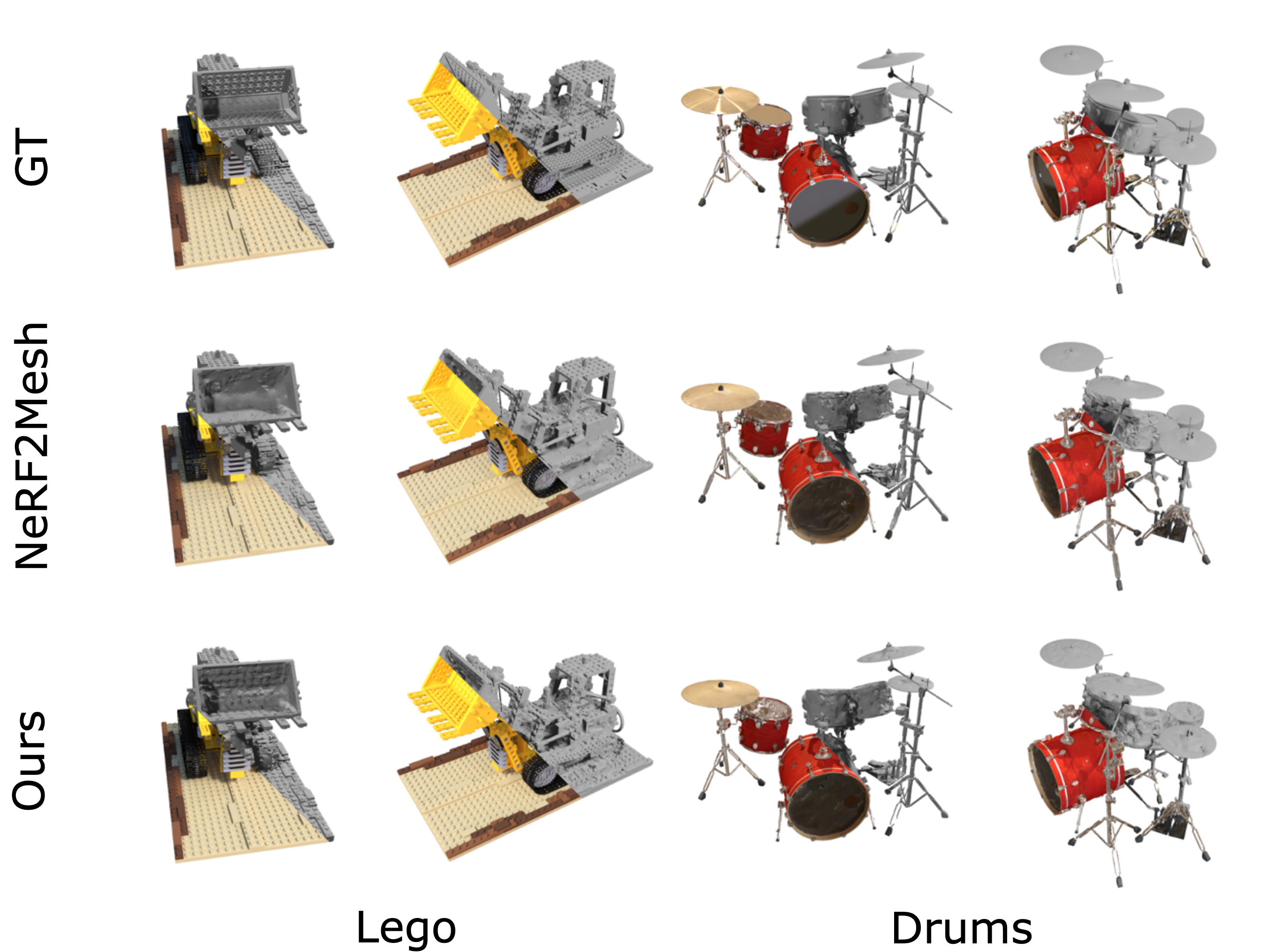}
  \vspace{-5pt}
  \caption{Visualization of our overall performance on the NeRF-synthetic dataset. Our method achieves highly competitive and robust performance in both mesh rendering quality and geometric quality.}
  \vspace{-10pt}
  \label{fig:overall}
\end{wrapfigure}

While these approaches have made notable progress, they typically rely on the given training set images. In this setting, supervision is restricted to a limited number of captured images, which is often insufficient to fully constrain geometry and appearance. Moreover, the contribution of each viewpoint for training is not only unequal but also changes dynamically as training progresses, so fixed viewpoints cannot provide the most effective guidance throughout the optimization process.

Our key insight is to go beyond the fixed captured images by exploiting the generative capacity of NeRF itself. NeRF can synthesize high-quality images from arbitrary camera poses, effectively serving as a source of additional pseudo-supervision. These NeRF-rendered views introduce complementary perspective information that is otherwise unavailable in the original dataset, thereby enriching the training signal and strengthening geometric constraints.

However, not all views are equally informative, and selecting unhelpful or redundant viewpoints may introduce noise or inefficiency. To address this, we propose an online viewpoint selection strategy based on the Upper Confidence Bound (UCB) algorithm \cite{ucb}. This reinforcement learning strategy dynamically balances exploration and exploitation, automatically identifying the most beneficial viewpoints at each stage of training. Combined with a geometry-aware reward, our method ensures that both real and NeRF-rendered images provide supervision from perspectives that maximally improve reconstruction.

Based on these insights, we propose \textbf{$R^2$-Mesh}, a reinforcement learning framework for high-quality mesh reconstruction. The framework jointly optimizes SDF geometry and view-dependent appearance under differentiable rendering supervision. Periodically, meshes are extracted from the evolving SDF and refined, allowing both geometry and connectivity to adapt dynamically over time. The overall performance is illustrated in Fig. \ref{fig:overall}.

In summary, the key contributions of our work are as follows:
\begin{itemize}
    \item We leverage NeRF-rendered images as additional pseudo-supervision, enriching the training signal with diverse and high-quality viewpoints that go beyond the original captures.
    
    \item We propose a UCB-based online viewpoint selection strategy with a geometry-aware reward, dynamically identifying the most informative views as training evolves.
    \item We present $R^2$-Mesh, a joint optimization framework of SDF and appearance, enabling progressive and topology-aware mesh refinement for high-fidelity reconstruction.
    
\end{itemize}

\section{Related Work}

\subsection{Mesh Reconstruction from NeRF}

NeRF \cite{nerf} and its variants \cite{tensorf,plenoxels,instantngp,dvgo} have significantly advanced 3D scene reconstruction by learning volumetric functions from multi-view images. Some works attempt to reconstruct meshes simultaneously during NeRF training. A representative example is MobileNeRF \cite{mobilenerf}, which directly generates polygonal meshes for real-time rendering. However, the produced meshes often lack smoothness since the topology is fixed at the initial stage and cannot adapt to complex geometry.

Recent studies \cite{neuralangelo,sparse-neus,nvdiffrec,unisurf,neus,neus2,volsdf} instead adopt Signed Distance Functions (SDFs) to represent geometry. Several of them leverage the ray-marching formulation of NeRF to optimize the SDF, followed by iso-surfacing algorithms such as Marching Cubes \cite{marching-cubes} or Dual Contouring \cite{dual-contouring} for mesh extraction. While effective, this post-processing often causes detail loss and surface artifacts, degrading mesh quality.

To mitigate these issues, some methods refine the extracted meshes. NVdiffrec \cite{nvdiffrec} employs differentiable mesh rendering to optimize the SDF, but random initialization may cause instabilities and floating artifacts. NeRF2Mesh \cite{nerf2mesh} refines vertex positions and face density based on rendering errors, but vertex connectivity remains fixed. Moreover, all these methods rely solely on the given training images, restricting supervision and overlooking that the training value of each viewpoint varies dynamically during optimization.

In contrast, our approach ensures a well-initialized representation, supports flexible adjustment of vertices and connectivity, and incorporates adaptive viewpoint selection to provide more effective supervision.

\begin{figure}[t]
\centering
\includegraphics[width=0.95\textwidth]{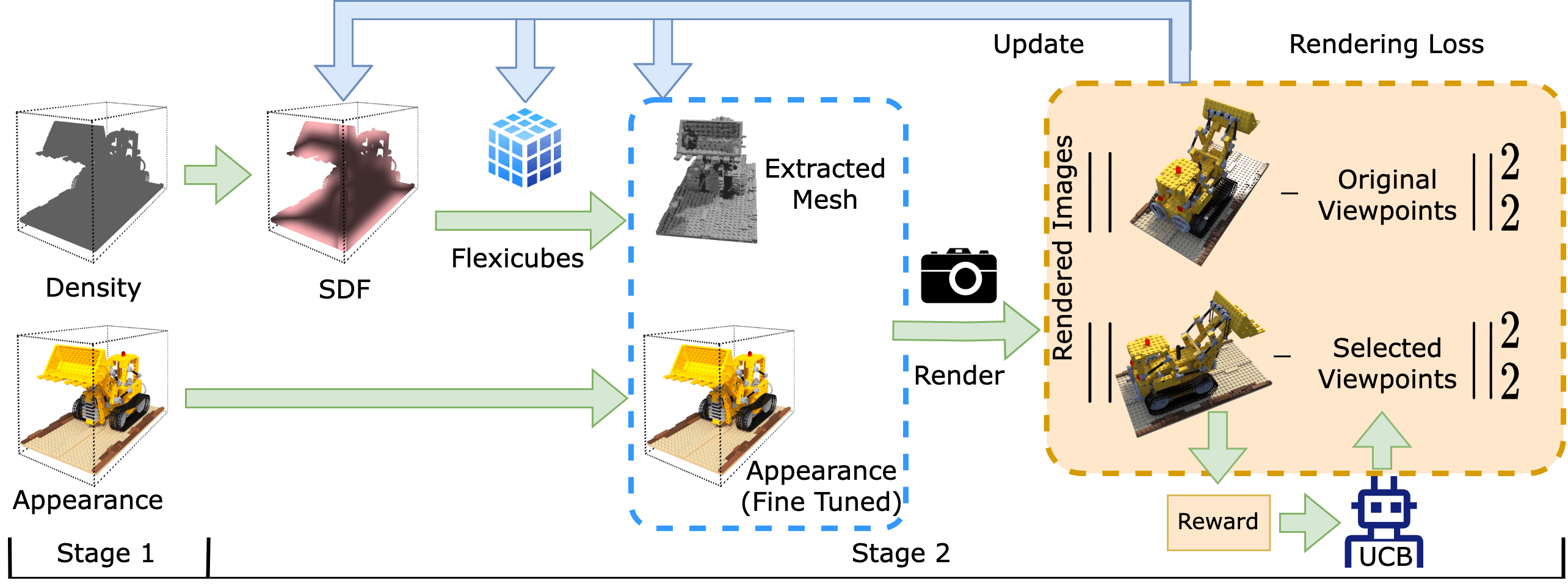} 
\caption{Our Framework, $R^2$-Mesh. In stage 1, we initialize the geometry and view-dependent appearance representation based on NeRF. This initial phase results in a coarse SDF grid and a set of candidate viewpoints rendered by NeRF model for enhancement in the subsequent stage. Then in stage 2, for each training iteration, we take two steps. We first select the optimal NeRF-rendered viewpoints based on UCB strategy to incorporate into the training dataset. We then simultaneously refine both the geometry and the appearance representation. After training is complete, we obtain the final mesh.}
\label{fig:pipeline}
\end{figure}

\subsection{Best Views Selection in 3d Scenes}

Selecting best views in 3D scenes is a critical issue in computer vision and 3D reconstruction, directly impacting the quality of final rendering and model reconstructions. Traditional approaches often rely on heuristic or predetermined viewpoint strategies, such as uniform sampling on a sphere or predefined camera trajectories. While straightforward, these strategies lack adaptability and often fail to capture the most informative views in scenes with complex geometry or strong occlusions.

Some recent works \cite{neu-nbv,activeneus,activenerf} in radiance fields incorporate uncertainty estimation to guide the selection process. By allocating more supervision to regions of high uncertainty, they improve rendering quality. However, these approaches are designed for volumetric ray sampling in radiance field training, and are less applicable to our focus on surface-based mesh reconstruction, where the supervision requirements differ significantly.

Additionally, several studies investigate next-best-view (NBV) planning \cite{krainin2011autonomous,mcgreavy2017next,naazare2022online}, primarily in robotic exploration of unknown environments. These methods aim to minimize the number of views required for complete coverage, whereas our objective is different: we aim to enhance mesh reconstruction quality by online selecting the most beneficial viewpoints from available candidates, based on their actual performance gains during training.

\section{Method}

\subsection{Framework}

In this section, we introduce our framework, \textbf{$R^2$-Mesh}, as illustrated in Fig. \ref{fig:pipeline}. We adopt a two-stage training process for mesh reconstruction. In stage 1, we utilize Instant-NGP \cite{instantngp} as the primary architecture to initialize the geometry and view-dependent appearance of the mesh (Section \ref{sec:stage1}). This initial phase results in a coarse SDF grid and a set of candidate viewpoints rendered by NeRF model for enhancement in the subsequent stage. Then in stage 2, for each training iteration, we take two steps. We first select the optimal combination of viewpoints based on UCB values to incorporate into the training dataset (Section \ref{sec:ucb}). We then simultaneously refine both the geometry and the appearance representation (Section \ref{sec:mesh_refine}). Upon completion of the training, we utilize xatlas \cite{xatlas} for UV unwrapping and export the resulting object mesh.

\subsection{Efficient 3d Scene Initialization (Stage 1)}
\label{sec:stage1}

In the first stage, we adopt the approach from \cite{nerf2mesh}, which builds on the Instant-NGP architecture to efficiently train a NeRF model for initializing 3D scene representation. The geometry is learned through a multi-resolutional density grid combined with a shallow Multilayer Perceptron (MLP). For appearance, the representation is decomposed into diffuse color and view-dependent specular components. The pixel color can be calculated by the following equations:

\begin{equation}
C(r) = \sum_{i=1}^N T_i (1 - \exp(-\sigma_i \Delta t_i)) \mathbf{c}_i, 
\label{eq:nerf_rendering}
\end{equation}
where \( C(r) \) is the accumulated color along ray \( r \), \( T_i \) is the transmittance from the camera to the \( i \)-th point, calculated as \( T_i = \exp\left(-\sum_{j=1}^{i-1} \sigma_j \Delta t_j\right) \). \( \sigma_i \) represents the density at the \( i \)-th point, \( \Delta t_i \) is the distance between the \( i \)-th and \( (i+1) \)-th sampled points, and \( \mathbf{c}_i \) is the combined diffuse and specular color at the \( i \)-th point.

To optimize the NeRF model, an mean squared error (MSE) loss is used to calculate the squared differences between the predicted and ground truth colors of the pixels:

\begin{equation}
L_{\text{MSE}} = \frac{1}{N} \sum_{i=1}^N (\hat{C}_i - C_i)^2, 
\label{eq:mse_loss}
\end{equation}
where \( \hat{C}_i \) is the predicted color and \( C_i \) is the ground truth color of the \( i \)-th pixel, with \( N \) representing the total number of pixels.

After training is complete, we extract a density grid at the specified resolution and convert it into an SDF grid. Density represents the amount of material at a point and is inherently positive, while SDF values indicate the shortest distance to the nearest surface, allowing them to be both positive and negative to distinguish between exterior and interior regions. Therefore, we set a threshold \(\theta\) to represent the iso-surface and convert density values into a coarse SDF grid using the following formula:
\begin{equation}
\text{SDF} =
\begin{cases} 
\frac{\sigma - \theta}{\max(\sigma) - \theta}, & \text{if } \sigma > \theta \\
\frac{\sigma - \theta}{\theta}, & \text{otherwise}
\end{cases},
\end{equation}
where \(\sigma\) represents the density value at a point, and \(\theta\) represents the iso-surface.

This transformation establishes our initial SDF grid, which we further refine in stage 2.

\begin{figure*}[t]
\centering
\includegraphics[width=0.95\textwidth]{mesh_blender.jpg} 
\caption{Mesh reconstruction quality on the NeRF-synthetic dataset. Our method produces significantly finer and more detailed mesh geometry compared to previous approaches.}
\vspace{-5pt}
\label{fig:mesh_geo}
\end{figure*}

\subsection{UCB-based Adaptive Viewpoint Enhancement (Stage 2)}
\label{sec:ucb}

After stage 1 converges, our goal in this stage is to refine the mesh by leveraging additional supervision beyond the initial training views. However, a key challenge lies in selecting the most informative viewpoints for continued refinement. As training progresses, both the model parameters and the rendering losses evolve, causing the effectiveness of each viewpoint to change over time. Relying on a fixed set of views may lead to redundant or suboptimal supervision. To address this, we propose an online reinforcement learning strategy based on the Upper Confidence Bound (UCB) algorithm, which adaptively selects viewpoints at each iteration by considering the current model state.

\begin{figure}[t]
\centering
\includegraphics[width=0.95\textwidth]{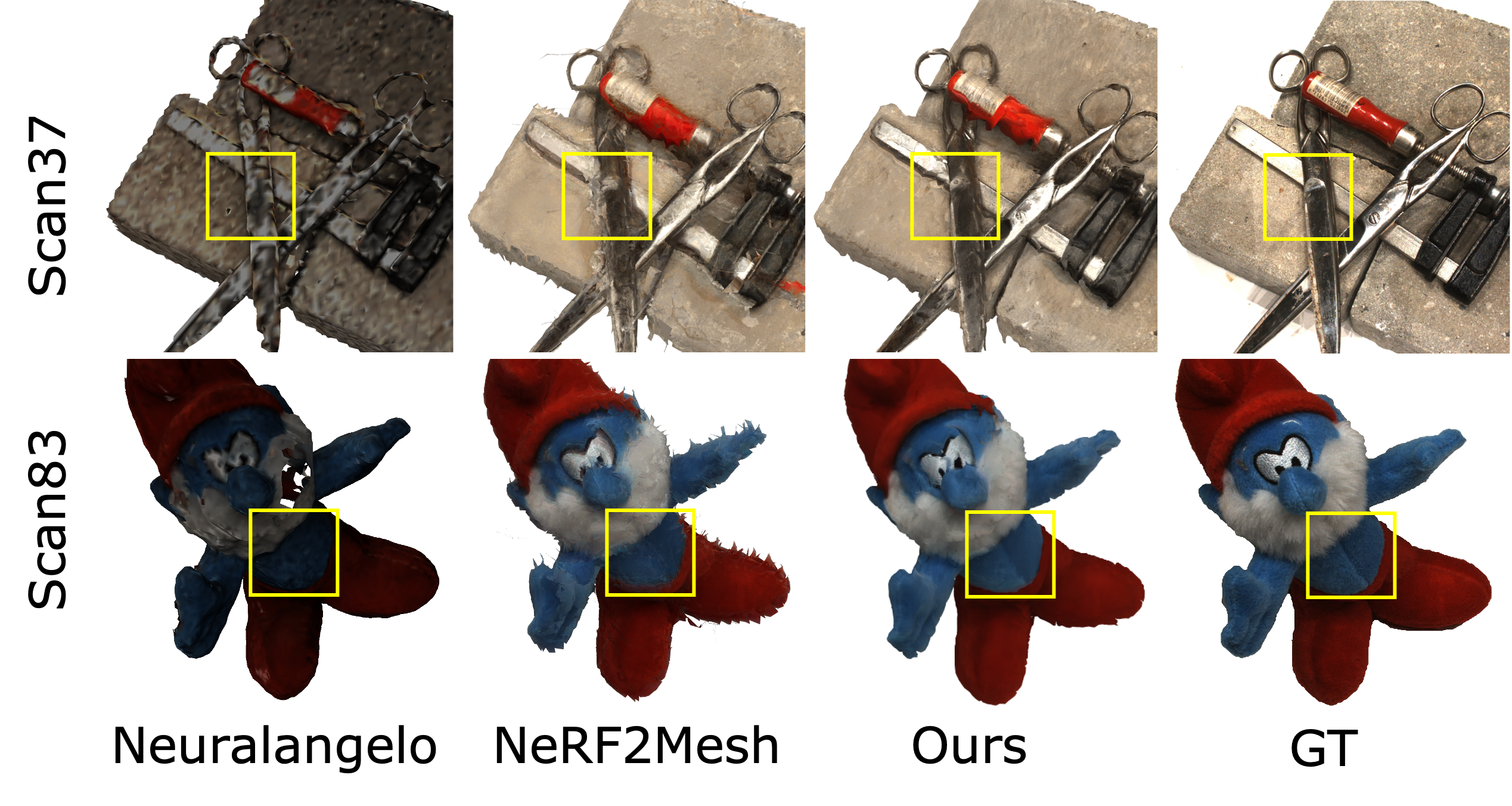} 
\caption{Visualization of rendering quality on the DTU dataset. Our method produces more accurate surface appearances compared to previous approaches.}
\label{fig:dtu_render_37_83}
\end{figure}

Many state-of-the-art reinforcement RL methods, such as DQN \cite{DQN} and PPO \cite{ppo}, introduce additional networks for inference, which results in significant time overhead. In contrast, our strategy using UCB does not require neural networks for inference, thereby reducing computational complexity and speeding up the decision-making process.

To support this selection process, we first generate a set of candidate viewpoints by rendering the scene from \( n \) uniformly distributed camera poses around a virtual sphere centered at the scene. Specifically, we define the candidate viewpoint set as \( V_{\text{NeRF}} = \{v_1, v_2, \ldots, v_n\} \), where each \( v_i \) corresponds to a camera located at a fixed radius \( r \) from the scene center, with elevation and azimuth angles sampled uniformly. These candidate views serve as the action space \( \mathcal{A} \) for the UCB algorithm.

At each iteration, we compute the UCB value for every candidate viewpoint (action) as:

\begin{equation}
\text{UCB}_a(t) = \hat{r}_a(t) + c \sqrt{\frac{2 \ln t}{N_a(t)}},
\end{equation}
where \( \hat{r}_a(t) \) represents the empirical mean reward for action \( a \) up to time \( t \), \( c \) is a constant controlling the degree of exploration, \(t \) represents the total number of steps \( t \), and \( N_a(t) \) denotes the number of times action \( a \) has been selected. Here, time \( t \) refers to the \( t \)-th iteration in the stage 2 training process. At initialization, the UCB value for each action is set to a large value, ensuring that every action is explored at least once.

For each training iteration, we randomly sample \(m\) viewpoints from the original training set \(V_{\text{train}}\) as ground truth supervision. In parallel, we evaluate the UCB values for all candidate viewpoints in the action space \(\mathcal{A} = V_{\text{NeRF}}\), and select the top-\(k\) actions with the highest UCB values. The corresponding viewpoints are treated as pseudo-ground-truth for this iteration. The mesh model is then trained using both the \(m\) ground truth images and the \(k\) NeRF-rendered pseudo ground truth images.

To guide the mesh refinement toward improved quality at novel views, we define the reward for each action \(a\) as the sum of two components:
\[
r_a = \alpha \, r_{\text{color}} + (1 - \alpha) \, r_{\text{geo}}, 
\quad \alpha \in [0,1].
\]

The \textbf{color reward} \(r_{\text{color}}\) evaluates how well the mesh captures the appearance at a given viewpoint. We use the MSE to enforce accurate pixel-level color alignment and the Learned Perceptual Image Patch Similarity (LPIPS) to capture perceptual differences that reflect high-level structural consistency. Formally,
\[
r_{\text{color}} = \text{MSE}(\text{Mesh}_a, \text{NeRF}_a) + \text{LPIPS}(\text{Mesh}_a, \text{NeRF}_a).
\]

To further leverage NeRF’s rich geometric information, we define a \textbf{geometry reward} \(r_{\text{geo}}\), which encourages alignment in the visible object regions between the mesh and the NeRF rendering. We compare their binary foreground indicators obtained by thresholding the rendered depth map. The reward is defined as:
\[
r_{\text{geo}} = \text{MSE}(\mathds{1}_{\text{Mesh}_a}, \mathds{1}_{\text{NeRF}_a}).
\]

Here, \(\mathds{1}_{\text{NeRF}_a}\) denotes the binary visibility map derived from NeRF’s depth rendering:
\[
\mathds{1}_{\text{NeRF}_a}(u, v) = \begin{cases}
1, & \text{if } D_{\text{NeRF}}(u, v) < \tau \\
0, & \text{otherwise}
\end{cases},
\]
where \(D_{\text{NeRF}}(u, v)\) is the NeRF-predicted depth at pixel \((u, v)\), and \(\tau\) is a fixed depth threshold to distinguish foreground regions.

By combining both appearance and geometric cues from NeRF, this reward formulation promotes improved rendering fidelity and reconstruction accuracy under both training and unseen viewpoints.

\subsection{Geometry and Appearance Refinement (Stage 2)}
\label{sec:mesh_refine}

After stage 1, we obtain a coarse representation of the scene in the form of an SDF grid and a view-dependent appearance field. While this initialization captures the overall geometry and appearance, it often lacks fine surface details and exhibit artifacts in poorly supervised regions. To achieve high-fidelity reconstruction, it is essential to further refine both geometry and appearance based on more precise surface supervision.

However, directly optimizing mesh vertex positions on a fixed topology limits the model’s ability to capture complex geometry. To address these issues, we refine both the SDF grid and the appearance representation based on surface rendering of the mesh. By directly optimizing the SDF, we enable continuous updates to both the mesh vertex positions and their connectivity, allowing the mesh geometry to flexibly adapt to complex surfaces as training progresses.

To ensure smooth gradient flow and enable topology-aware refinement, we adopt FlexiCubes \cite{flexicubes}, which augments each SDF grid vertex with learnable deformation and weight parameters. These attributes allow the extracted mesh to deform continuously and adjust its connectivity during optimization.

At each training iteration, we take two steps. We first select the optimal viewpoints to enhance the training dataset based on our UCB strategy. We then extract the mesh from the SDF using FlexiCubes and render it with nvdiffrast \cite{nvdiffrec}, enabling end-to-end optimization of both geometry and appearance through differentiable rendering.

\begin{wrapfigure}{r}{0.45\textwidth}
  \centering
  \vspace{-2pt}  
  \includegraphics[width=0.43\textwidth]{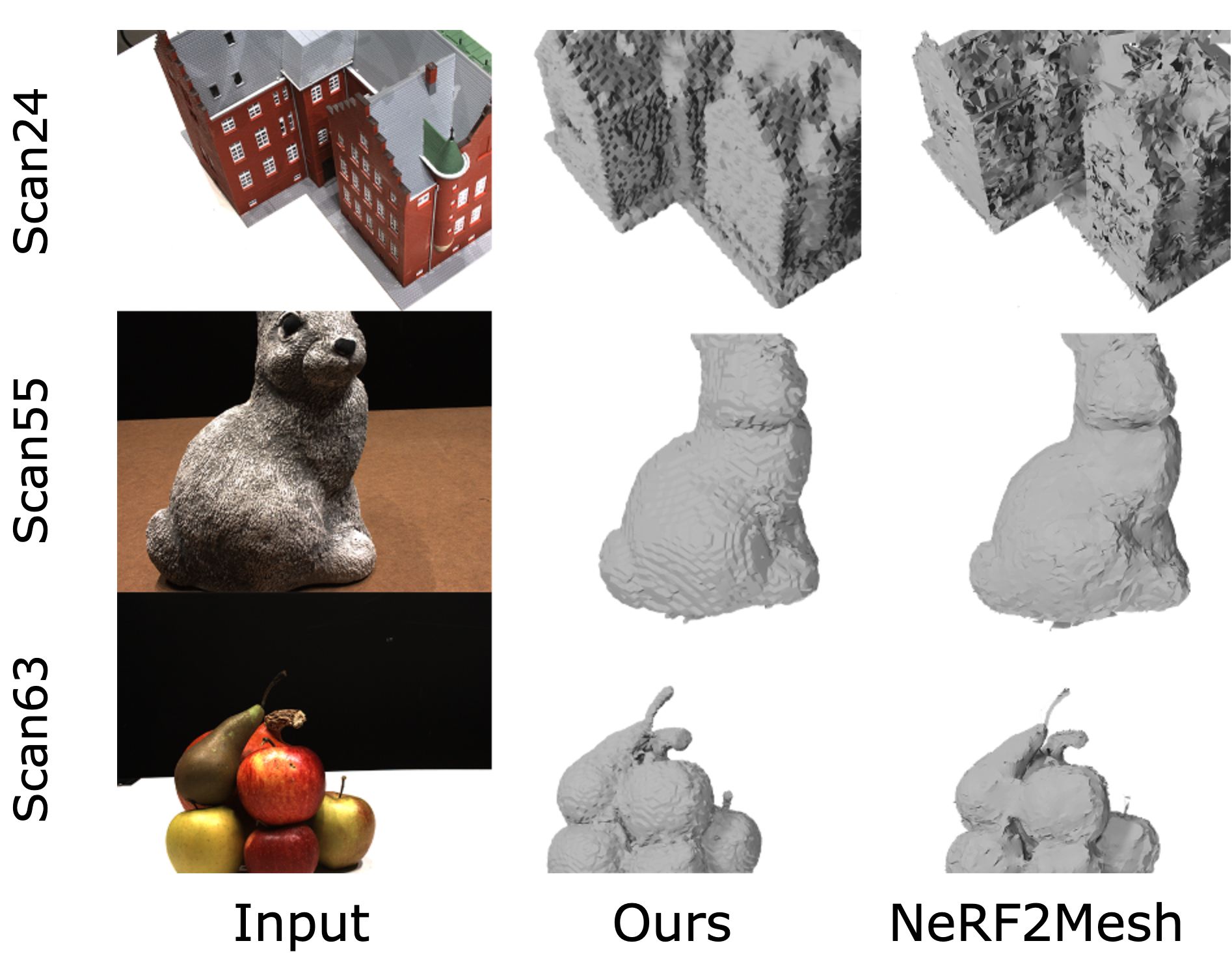}
  \vspace{-5pt}
  \caption{Mesh reconstruction quality on the DTU dataset. Our method delivers better performance than previous methods.}
  \label{fig:mesh_dtu}
\end{wrapfigure}

To optimize the parameters, the overall loss functions are:  
\begin{equation}
    L = \lambda_{\text{color}}L_{\text{color}}(x) + \lambda_{\text{TV}} L_{\text{TV}} + \lambda_{\text{dev}} L_{\text{dev}},
\end{equation}
where \( L_{\text{color}}(x) = \sqrt{(x - x^*)^2 + \epsilon^2} \) is the Charbonnier loss \cite{mipnerf}, with \(x\) as the predicted value, \(x^*\) as the ground truth, and \(\epsilon\) as a small constant to improve robustness to outliers. \(L_{\text{TV}}\) is the Total Variation (TV) regularization term \cite{tensorf} applied to the SDF grid to reduce floaters and enhance smoothness, while \(L_{\text{dev}}\) is the FlexiCubes regularizer used to suppress mesh artifacts. Upon completing the training process, we employ xatlas to perform UV unwrapping on the generated mesh and subsequently export the final object mesh.

\section{Experiments}

\subsection{Implementation Details}

\subsubsection{Datasets}

To evaluate our approach and baseline approaches, we use 15 scenes from the real-world DTU dataset \cite{dtu} and all scenes from the NeRF-synthetic \cite{nerf} dataset to assess both the reconstructed mesh quality and the rendering quality. The DTU dataset provides 49 or 64 images per scene captured from fixed viewpoints in a controlled indoor setup, while the NeRF-synthetic dataset contains 100 training images per scene rendered from synthetic objects along a predefined spherical trajectory. Experimental results demonstrate that our approach achieves excellent rendering and mesh reconstruction outcomes across most real-world and synthetic scenes.

\subsubsection{Training Details}

In the first stage, we train the NeRF model for 30,000 steps. At the end of the training, we extract an SDF grid at a resolution of \(128^3\). For stage 2, we train the model for 90,000 steps to ensure that both the UCB and the model converge. The optimizer used is Adam, with the learning rates for both the FlexiCubes parameters and the SDF set to \(1 \times 10^{-4}\). The experiments are all conducted on a high-performance computing platform equipped with a GeForce RTX 3090 GPU.

\begin{figure}[t]
  \centering
  \begin{minipage}[t]{0.48\textwidth}
    \centering
    \vspace{-80pt}
    \includegraphics[width=\linewidth]{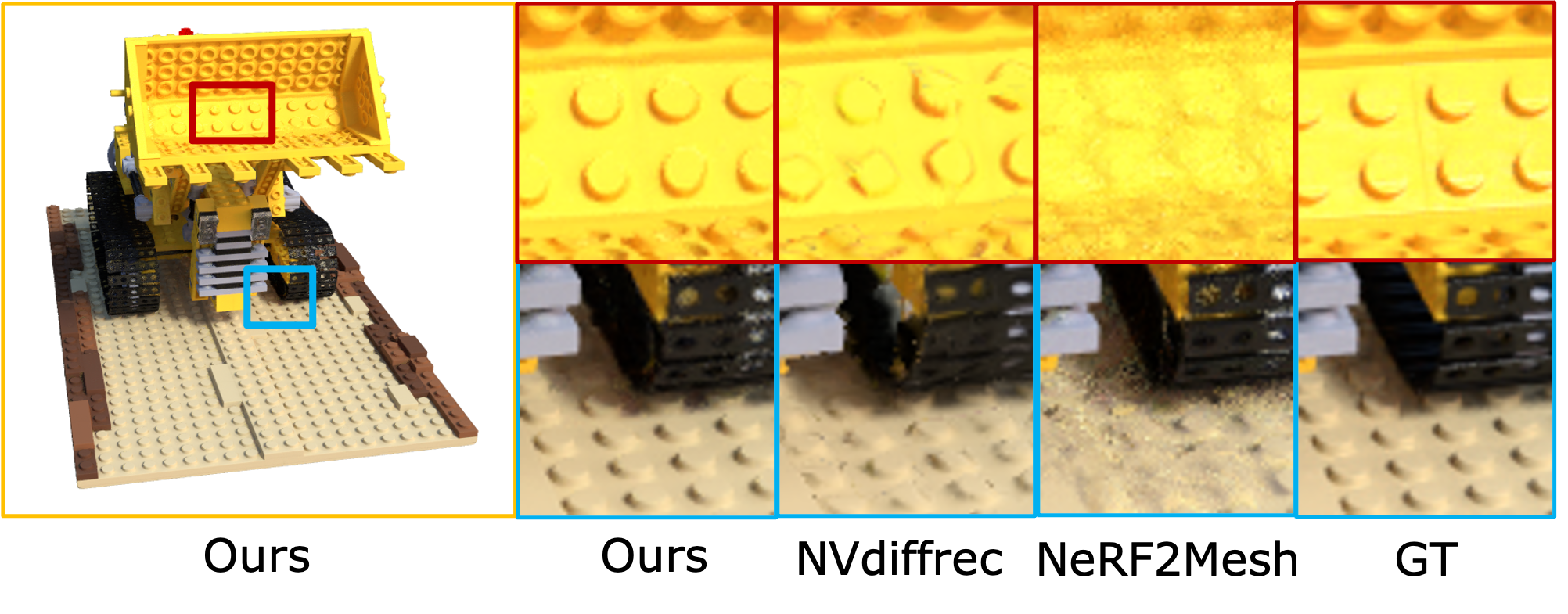}
    \vspace{-18pt}
    \caption{Visualization of rendering quality on the NeRF-synthetic dataset. Our method can render more detailed geometry.}
    \label{fig:fig_render}
  \end{minipage}
  \hfill
  \begin{minipage}[t]{0.48\textwidth}
    \centering
    \includegraphics[width=\linewidth]{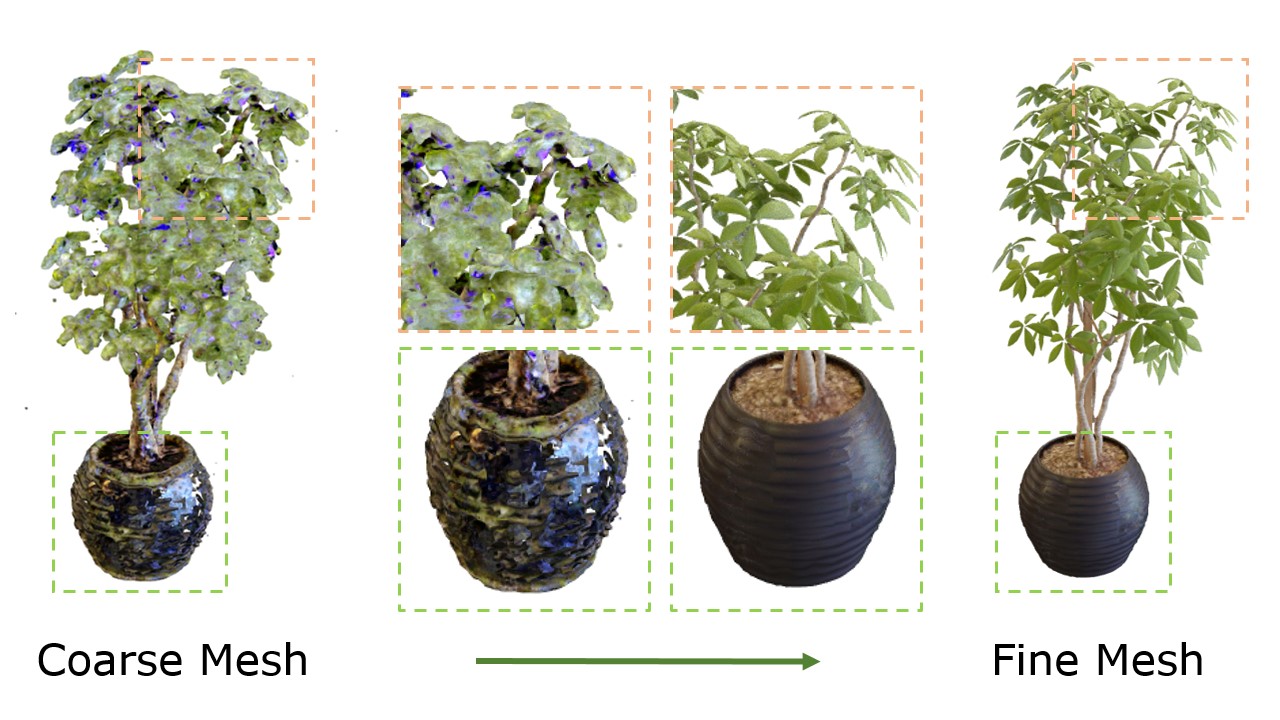}
    \caption{Comparison between coarse and fine mesh.}
    \label{fig:fig_coarse2fine}
  \end{minipage}
\end{figure}

\subsection{Evaluation}

\subsubsection{Mesh Quality}


To evaluate the geometric accuracy of our mesh reconstructions, we utilize the Chamfer Distance (CD) metric which quantifies the average closest point distance between the vertices of the reconstructed mesh and the ground truth. We establish MobileNeRF\cite{mobilenerf}, NVdiffrec\cite{nvdiffrec}, NeuS2\cite{neus2}, Neuralangelo\cite{neuralangelo}, NeRFMeshing\cite{nerfmeshing} and NeRF2Mesh\cite{nerf2mesh} as baselines for our experiments and find that our results significantly outperform the previous works. The detailed outcomes of our mesh reconstruction are presented in Table \ref{tab:table_chamber_distance_nerf}, Table \ref{tab:table_chamber_distance_dtu}, Fig. \ref{fig:mesh_geo} and Fig. \ref{fig:mesh_dtu}.

\begin{table*}[th]
\caption{Chamfer Distance (\(\downarrow\)) on the NeRF-synthetic dataset compared to the ground truth geometry. Our method achieves competitive results compared to other methods. Red and orange indicate the first and second best performing algorithms for each scene.}
\begin{center}
\small
\begin{tabular}{c|cccccccc|c}
  \toprule
  Method & Lego & Hotdog & Chair & Drums & Ficus & Mic & Ship & Materials & Mean \\
  \midrule
  NVdiffrec 
  & 1.9 
  & \cellcolor{red!20}\textbf{3.1}
  & 9.4 
  & 3.4 
  & \cellcolor{orange!20}0.6 
  & 1.0 
  & 38.4 
  & \cellcolor{orange!20}1.2 
  & 7.38  \\

  NeuS2 
  & 2.0 
  & \cellcolor{orange!20}3.4 
  & 4.1 
  & \cellcolor{orange!20}1.8 
  & \cellcolor{orange!20}0.6 
  & \cellcolor{orange!20}0.9 
  & \cellcolor{orange!20}19.6 
  & 1.3 
  & 4.22 \\

  NeRF2Mesh 
  & 2.1 
  & 3.7 
  & \cellcolor{red!20}\textbf{1.0} 
  & 2.5 
  & \cellcolor{red!20}\textbf{0.4} 
  & 1.2 
  & 35.9 
  & \cellcolor{orange!20}1.2 
  & 6.00 \\

  NeRFMeshing
  & \cellcolor{red!20}\textbf{1.5}
  & 4.2
  & \cellcolor{red!20}\textbf{1.0} 
  & \cellcolor{orange!20}1.8
  & 0.8
  & 1.3 
  & \cellcolor{red!20}10.4
  & 1.4
  & \cellcolor{orange!20}2.80 \\

  Ours 
  & \cellcolor{orange!20}1.6  
  & \cellcolor{red!20}\textbf{3.1} 
  & \cellcolor{orange!20}3.3 
  & \cellcolor{red!20}\textbf{1.6} 
  & \cellcolor{red!20}\textbf{0.4} 
  & \cellcolor{red!20}\textbf{0.8} 
  & \cellcolor{red!20}\textbf{10.4} 
  & \cellcolor{red!20}\textbf{0.5} 
  & \cellcolor{red!20}\textbf{2.71} \\
  \bottomrule
\end{tabular}
\vspace{-30pt}
\label{tab:table_chamber_distance_nerf}
\end{center}
\end{table*}

\begin{table*}[th]
\begin{center}
\caption{Chamfer Distance (\(\downarrow\)) on the DTU dataset compared to the ground truth geometry. Our method achieves competitive results compared to other methods. Red and orange indicate the first and second best performing algorithms for each scene.}
\resizebox{\textwidth}{!}{
\begin{tabular}{c|ccccccccccccccc|c}
  \toprule
  \multirow{2}{*}{Method} & \multicolumn{15}{c|}{Scan} & \multirow{2}{*}{Mean} \\
  & 24 & 37 & 40 & 55 & 63 & 65 & 69 & 83 & 97 & 105 & 106 & 110 & 114 & 118 & 122 & \\
  \midrule
  NeuS2 
  & \cellcolor{orange!20}0.56 
  & \cellcolor{orange!20}0.76 
  & \cellcolor{orange!20}0.49 
  & \cellcolor{orange!20}0.37 
  & \cellcolor{red!20}\textbf{0.92} 
  & 0.71 
  & \cellcolor{red!20}\textbf{0.76} 
  & \cellcolor{orange!20}1.22 
  & 1.08 
  & \cellcolor{orange!20}0.63
  & \cellcolor{orange!20}0.59 
  & \cellcolor{red!20}\textbf{0.89} 
  & \cellcolor{red!20}\textbf{0.40} 
  & \cellcolor{orange!20}0.48 
  & 0.55 
  & \cellcolor{orange!20}0.69 \\

  NeRF2Mesh 
  & 0.59 
  & 0.89 
  & 0.51 
  & 0.45 
  & 1.03 
  & \cellcolor{red!20}\textbf{0.65} 
  & 0.83 
  & 1.27 
  & \cellcolor{orange!20}1.03 
  & \cellcolor{red!20}\textbf{0.62} 
  & 0.72 
  & \cellcolor{orange!20}1.05 
  & 0.66 
  & 0.72 
  & \cellcolor{orange!20}0.49 
  & 0.77 \\

  Ours 
  & \cellcolor{red!20}\textbf{0.51} 
  & \cellcolor{red!20}\textbf{0.65} 
  & \cellcolor{red!20}\textbf{0.38} 
  & \cellcolor{red!20}\textbf{0.32} 
  & \cellcolor{orange!20}0.96
  & \cellcolor{orange!20}0.68 
  & \cellcolor{orange!20}0.79 
  & \cellcolor{red!20}\textbf{1.12} 
  & \cellcolor{red!20}\textbf{0.82} 
  & 0.87 
  & \cellcolor{red!20}\textbf{0.57} 
  & 1.08 
  & \cellcolor{orange!20}0.44 
  & \cellcolor{red!20}\textbf{0.41} 
  & \cellcolor{red!20}\textbf{0.44} 
  & \cellcolor{red!20}\textbf{0.67} \\
  \bottomrule
\end{tabular}
}
\label{tab:table_chamber_distance_dtu}
\end{center}
\end{table*}

\subsubsection{Rendering Quality}

We employ PSNR, SSIM (Structural Similarity Index Measure), and LPIPS metrics to evaluate the rendering quality of our mesh reconstruction. Compared to baselines, our methodology demonstrated superior performance in the majority of the tested scenes. Note that NeRFMeshing only reports PSNR on its benchmarks, NVdiffrec results are not reported on the DTU dataset, as the official release did not include the dataset reader and the MLP-based geometry representation required for evaluation. The detailed results of this comparative analysis are presented in Table \ref{tab:table_render}, Fig. \ref{fig:dtu_render_37_83} and Fig. \ref{fig:fig_render}. These metrics highlight our approach's advancements in delivering high-fidelity reconstructions across various complex scenarios.

\begin{table}[h]
\caption{Rendering quality comparisons on the NeRF-synthetic (SYN) and the DTU dataset. Compared to Neuralangelo, NVdiffrec, NeRFMeshing and NeRF2Mesh, our method achieves better rendering quality. Red and orange indicate the first and second best performing algorithms for each scene. (VE: Viewpoint Enhancement) }
\begin{center}
\resizebox{0.6\linewidth}{!}{
\begin{tabular}{c|cc|cc|cc}
    \toprule
    \multirow{2}{*}{Method} & \multicolumn{2}{c|}{PSNR ↑} & \multicolumn{2}{c|}{SSIM ↑} & \multicolumn{2}{c}{LPIPS ↓} \\ 
    & SYN & DTU & SYN & DTU & SYN & DTU \\ 
    \midrule
    Neuralangelo       
    & 27.12 
    & 21.98 
    & 0.91 
    & \cellcolor{orange!20}0.88 
    & 0.091 
    & 0.25 \\

    NeRFMeshing      
    & 27.31
    & -
    & - 
    & - 
    & -
    & - \\

    NVdiffrec       
    & 28.76 
    & - 
    & \cellcolor{orange!20}0.93 
    & -
    & 0.049 
    & - \\

    NeRF2Mesh        
    & 29.11 
    & 22.46 
    & \cellcolor{orange!20}0.93 
    & \cellcolor{red!20}\textbf{0.90} 
    & 0.080 
    & \cellcolor{orange!20}0.21 \\

    Ours (w/o VE)    
    & \cellcolor{orange!20}29.26 
    & \cellcolor{orange!20}22.84 
    & \cellcolor{red!20}\textbf{0.94} 
    & \cellcolor{red!20}\textbf{0.90} 
    & \cellcolor{orange!20}0.048
    & \cellcolor{red!20}\textbf{0.13} \\

    Ours             
    & \cellcolor{red!20}\textbf{29.55} 
    & \cellcolor{red!20}\textbf{23.20} 
    & \cellcolor{red!20}\textbf{0.94} 
    & \cellcolor{red!20}\textbf{0.90} 
    & \cellcolor{red!20}\textbf{0.046} 
    & \cellcolor{red!20}\textbf{0.13} \\
    \bottomrule
\end{tabular}
}
\label{tab:table_render}
\vspace{-30pt}
\end{center}
\end{table}

\begin{wrapfigure}{r}{0.6\textwidth}
  \centering
  \vspace{-20pt}  
  \includegraphics[width=0.58\textwidth]{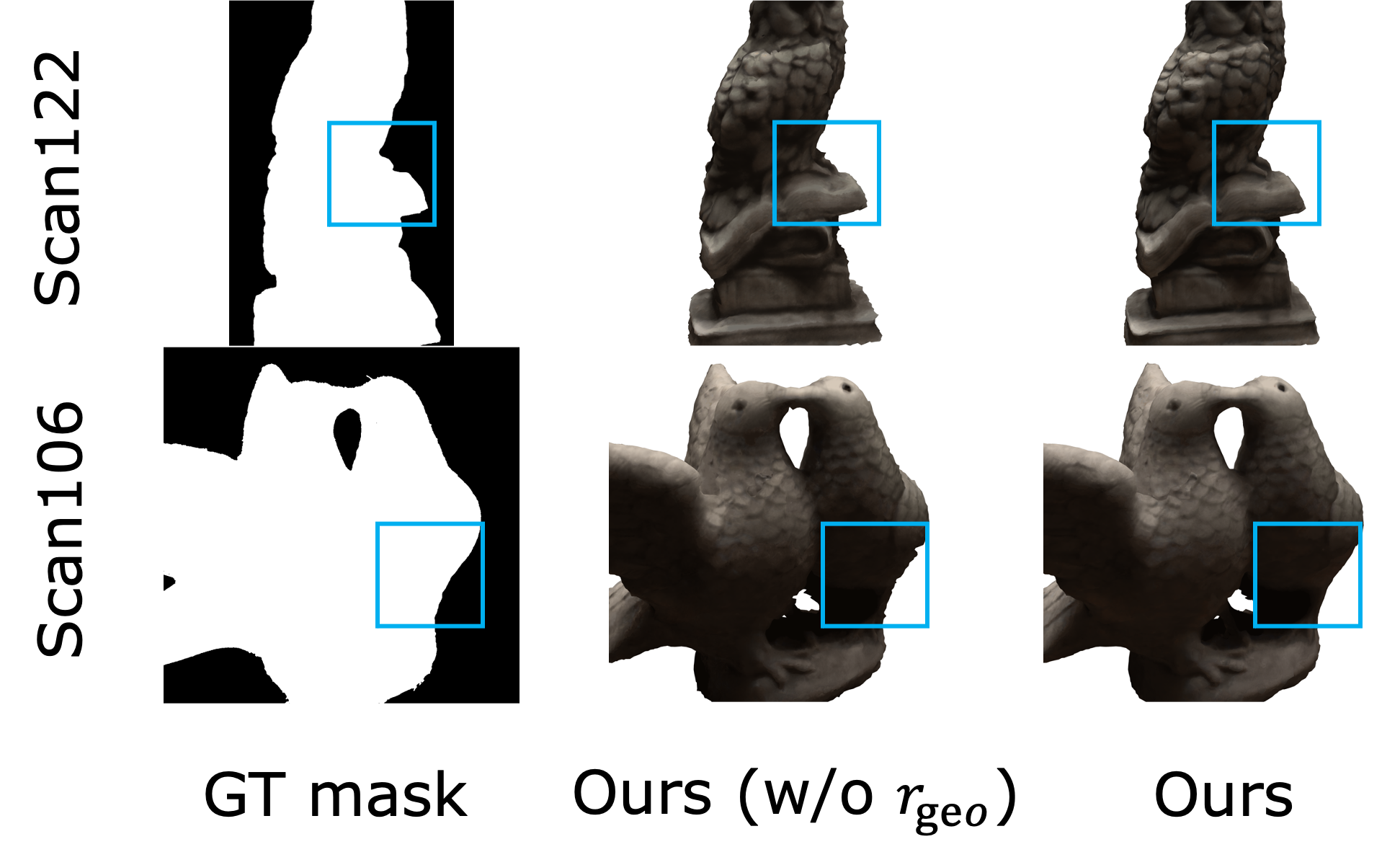}
  \vspace{-5pt}
  \caption{Ablation study on geometric reward. With the geo reward, the mesh shows cleaner boundaries and fewer artifacts.}
  \vspace{-20pt}
  \label{fig:ablation_geo}
\end{wrapfigure}

\subsection{Ablation Studies}

We conduct an ablation study to evaluate the effectiveness of our mesh refinement method, the online UCB-based adaptive viewpoint enhancement strategy, and the geometric reward \(r_{\text{geo}}\). We compare the performance of the full model against variants that exclude mesh refinement (RF), viewpoint enhancement (VE), and geometric reward. We also assess the effectiveness of the UCB strategy by comparing it to random and greedy viewpoint selection baselines. In the greedy strategy, we select viewpoints nearest to the one with the highest loss at each training iteration.

\begin{table}[t]
    \begin{minipage}{0.47\linewidth}
    \caption{Ablation study of our design choices. The results show that both the mesh refinement (RF) and viewpoint enhancement (VE) improve the overall performances.}
        \centering
        \resizebox{\linewidth}{!}{
        \begin{tabular}{cc|cc|cc|cc}
            \toprule
            \multirow{2}{*}{VE} & \multirow{2}{*}{RF} & \multicolumn{2}{c|}{PSNR ↑} & \multicolumn{2}{c|}{SSIM ↑} & \multicolumn{2}{c}{LPIPS ↓} \\ 
              & & SYN & DTU & SYN & DTU & SYN & DTU \\ 
            \midrule
            \ding{51} & \ding{51} & \textbf{29.55} & \textbf{23.20}    & \textbf{0.94} & \textbf{0.90}    & \textbf{0.046} & \textbf{0.13}    \\
            \ding{55} & \ding{51} & 29.26 & 22.84    & 0.94 & 0.90    & 0.048 & 0.13    \\
            \ding{55} & \ding{55} & 15.43 & 16.10 & 0.72 & 0.65 & 0.27 & 0.42 \\
            \bottomrule
        \end{tabular}
        }
        
        \label{tab:table_ablation}
    \end{minipage}
    \hfill
    \begin{minipage}{0.49\linewidth}
    \caption{Ablation study of our viewpoint enhancement strategies. Our UCB strategy outperforms both random and greedy strategies.}
        \centering
        \resizebox{\linewidth}{!}{
        \begin{tabular}{c|cc|cc|cc}
            \toprule
            \multirow{2}{*}{Strategy} & \multicolumn{2}{c|}{PSNR ↑} & \multicolumn{2}{c|}{SSIM ↑} & \multicolumn{2}{c}{LPIPS ↓} \\ 
              & SYN & DTU & SYN & DTU & SYN & DTU \\ 
            \midrule
            Ours & \textbf{29.55} & \textbf{23.20}    & \textbf{0.94} & \textbf{0.90}    & \textbf{0.046} & \textbf{0.13}    \\
            Ours (w/o \(r_{\text{geo}}\)) & 29.47 & 23.07    & 0.94 & 0.90    & 0.046 & 0.13    \\
            Greedy & 29.18 & 22.87 & 0.93 & 0.90  & 0.052 & 0.15 \\
            Random & 29.38 &  23.08 & 0.94 & 0.90  & 0.047 & 0.13 \\
            \bottomrule
        \end{tabular}
        }
        
        \label{tab:table_ucb}
        
    \end{minipage}
    \vspace{-10pt}
\end{table}

As shown in Table \ref{tab:table_ablation}, when viewpoint enhancement is removed, the rendering quality decreases, primarily because the model lacks the ability to incorporate diverse perspectives that contribute to more detailed and accurate reconstructions. When mesh refinement is further removed, there is a significant decline in the rendering quality of the mesh. The comparison between fine mesh and coarse mesh is shown in Fig. \ref{fig:fig_coarse2fine}. Additionally, removing the geometric reward \(r_{\text{geo}}\) results in increased artifacts along object boundaries, as shown in Fig. \ref{fig:ablation_geo}.

Table \ref{tab:table_ucb} shows the UCB strategy outperforms both random and greedy selection methods. This is because the greedy approach tends to overfit to the currently known best viewpoints, potentially missing out on more informative perspectives. In contrast, the UCB method effectively balances exploration and exploitation, allowing for the online selection of the optimal viewpoints.

\section{Conclusion}

In summary, we present $R^2$-Mesh, a framework that leverages an online reinforcement learning strategy based on the UCB algorithm to adaptively select the most informative viewpoints for mesh reconstruction. Starting from a NeRF-trained coarse SDF and appearance field, our method refines geometry and appearance through differentiable mesh extraction and rendering. By combining NeRF-rendered pseudo-supervision with UCB-guided viewpoint selection, our method achieves substantial improvements in geometric fidelity and rendering quality over existing approaches. 

\begin{credits}
\subsubsection{\ackname} This work was sponsored by the NSFC grant(62431017). We gratefully acknowledge the support of State Key Laboratory of Media Convergence Production Technology and Systems, Key Laboratory of Intelligent Press Media Technology. Xinggong Zhang is the corresponding author.
\end{credits}

%
%
%
%

\bibliographystyle{splncs04}
\bibliography{mybib}
\end{document}